%% file: root.tex
\documentclass[letterpaper, 10pt, conference]{ieeeconf}
\IEEEoverridecommandlockouts                   % needed for \thanks
\usepackage{graphicx}
\usepackage{amsmath,amssymb,amsfonts}
\usepackage{booktabs}
\usepackage{multirow}
\usepackage{float}
\usepackage[caption=false,font=footnotesize]{subfig}
\usepackage[colorlinks=true, allcolors=blue]{hyperref}
\makeatletter
\providecommand{\@extra@b@citeb}{}
\def\@citex[#1]#2{\leavevmode
  \let\@citea\@empty
  \@cite{\@for\@citeb:=#2\do
    {\@citea\def\@citea{], [}%
     \edef\@citeb{\expandafter\@firstofone\@citeb\@empty}%
     \if@filesw\immediate\write\@auxout{\string\citation{\@citeb}}\fi
     \@ifundefined{b@\@citeb}{\mbox{\reset@font\bfseries ?}%
       \G@refundefinedtrue
       \@latex@warning{Citation `\@citeb' on page \thepage\space undefined}}%
     {\hyperlink{cite.\@citeb\@extra@b@citeb}%
       {\csname b@\@citeb\endcsname}}}}{#1}}
\let\Hy@origbibitem\bibitem
\renewcommand{\bibitem}[1]{\Hy@origbibitem{#1}\hypertarget{cite.#1}{}}
\makeatother
\usepackage{xcolor}
\usepackage{siunitx}
\usepackage{dblfloatfix}

\newcommand{\method}{DeltaForge}
\newcommand{\minade}{minADE}

\title{\LARGE \bf
\method{}: A Post-hoc Mixture-of-Experts VLA for Multi-Domain Autonomous Driving
}

\author{
Bingzhang Wang$^{1,2}$,
Mo Poorsartep$^{1}$,
and Yasaman Esfandiari$^{1}$%
\thanks{$^{1}$NVIDIA Corporation, Santa Clara, CA, USA.}%
\thanks{$^{2}$University of Washington, Seattle, WA, USA.}%
}

\begin{document}

\maketitle
\thispagestyle{empty}
\pagestyle{empty}

% -----------------------------------------------------------------------------
\begin{abstract}
Fleets of driving vision--language--action (VLA) models are developed one
domain at a time: each operator finetunes a shared base model on private data,
producing specialized experts that are costly to consolidate into a single model. Weight
merging loses accuracy as heterogeneous domains accumulate, and dense
mixture-of-experts routing is prohibitively expensive to combine them. We
present \method{}, to our knowledge the first post-hoc mixture-of-experts VLA
for driving, which consolidates $N$ independently finetuned per-domain models
into a single multi-domain model from their weights and a small proxy set
alone. 
% Motivated by the low-rank and redundant structure of finetuning updates,
% its post-hoc expert factorization applies a joint singular value decomposition (SVD) across all domain deltas
% to extract a frozen skill basis shared across domains together with a small
% per-domain low-rank core, so that routing operates on the compact cores rather
% than dense expert blocks.
When different operators finetune the same base model, their weight updates end up sharing a lot of structure. We exploit this by decomposing all the updates together using singular value decomposition (SVD), splitting each one into a shared part (the same across all domains) and a tiny domain-specific part. The router then only needs to work with those tiny domain-specific pieces, not the full expert weights, which is what keeps the approach efficient.
Then two cooperating clip-level blind routers, one over the
VLM reasoning experts and one over the flow-matching dynamics experts, are
trained from soft labels on the proxy dataset. 
Across six diverse domains
spanning urban, off-road, and highway driving on cars, trucks, and all-terrain vehicles (ATVs),
\method{} approaches the accuracy of the per-domain experts and surpasses every
merging, routing, centralized, and federated baseline at close to single-model
parameter and compute cost. 
% A new domain opts in by projecting its delta onto
% the frozen basis without retraining, and the composed model exhibits
% compositional generalization to unseen domains.
Adding a new domain requires only projecting its weight update onto the shared basis with no end-to-end retraining needed. The resulting model also generalizes to domain combinations it was never explicitly trained on.
\end{abstract}

% -----------------------------------------------------------------------------
\input{sections/01_introduction}
\input{sections/02_related_work}

\input{sections/03_method}
\input{sections/04_experiments}
\input{sections/05_conclusion}

% -----------------------------------------------------------------------------
\bibliographystyle{IEEEtran}
\bibliography{refs}

\end{document}

%% file: sections/01_introduction.tex
\section{Introduction}
\label{sec:intro}

% =============================================================================
% Paragraph plan:  P1 vision | P2-P5 challenges (data availability,
%                  heterogeneity, efficiency, flexibility) | P6 research question
%                  P7 approach | P8 contributions
% The original draft is kept verbatim in comments below each slot.
% =============================================================================

% ---- P1: Vision -------------------------------------------------------------
VLA models are emerging as foundation policies for
end-to-end autonomous driving: a single network perceives the scene, reasons
about it in language space, and outputs a trajectory. Real deployments,
however, are fragmented. Driving systems operate across very different environments, vehicles, and
regions, so each operator finetunes the shared base model on its own data,
% and the field accumulates 
leaving the field with a growing collection of specialized domain models rather
than one generalizable foundation model. This paper asks
how to consolidate these domain models back into a single multi-domain VLA, a
goal blocked by four coupled obstacles: data availability, domain
heterogeneity, efficiency, and flexibility.
% --- original draft ----------------------------------------------------------
% Towards one generlizable driving VLA fleet models across heterogeous use case
% domains is a ultmiate goal of having a driving foundation model. While, this
% goal is blocked by multiple factors such as data privacy concerns, domain
% heterogeneity, model efficiency, flexibility of adding new domains.
% -----------------------------------------------------------------------------

% ---- P2: Challenge C1 — data availability -----------------------------------
\textbf{Data availability.}
In realistic deployments, this domain data is often privacy-sensitive or
regulated and cannot be
pooled for centralized training: only model weights leave a domain. Federated
learning \cite{mcmahan2017communication} respects this constraint, but it
requires all clients to
train synchronously, and under strongly heterogeneous domains its averaged
global model suffers from gradient divergence. What is needed instead is a
framework that aggregates \emph{already-finetuned weights} into one
multi-domain VLA without access to the underlying datasets.
% --- original draft ----------------------------------------------------------
% privacy/data availability: in the real use cases, each entity finetunes their
% models on their own individual domain data. These data may be sensitive and
% not available for sharing data across domains for pooled centralized
% training. For example, a china company has their own data that are privacy
% sensitive and has regulation not to share datasets. A framework needs to be
% devised to aggregate their trained weights without sharing entire in-domain
% data, in order to have a generalizable, multi-domain capable VLA models. FL
% (cite) can be a solution to this limit, while they require synchroinzed
% training within multi-domain clients avaialbe for training. Also, the
% aggregated centralized model may suffer gradient divergence under
% hetergenerous domain seetings as decribed in the later point.
% -----------------------------------------------------------------------------

% ---- P3: Challenge C2 — domain heterogeneity --------------------------------
\textbf{Domain heterogeneity.}
Driving domains differ along three axes: environment (urban, highway,
off-road, parking, and different regions such as China, Japan, or the US),
vehicle embodiment and dynamics (passenger cars, trucks, ATVs,
with varying camera counts and placements), and regulation (e.g., left- versus
right-hand traffic). Unlike text-only LLM tasks, a driving VLA must map such
heterogeneous visual and kinematic inputs to kinematically feasible
trajectories, a multi-modal setting in which mixture-of-experts (MoE)
modeling is both more necessary and far less studied. A mixture of domain
experts is the natural remedy, but classical MoE trains its sparse
architecture from scratch on pooled data, which the data constraint above
forbids.
% --- original draft ----------------------------------------------------------
% heterogeneity: The driving domains can be highly heterogeneous, with
% in-distribution and out-of-distribution domains compared to pretrained base
% data. Unlike language tasks for LLMs, the modality of language is very not
% diverse. For AV VLA models MoE is challenging and lack of study, it has more
% complex model architecture to learn from visual, dynamic features prejecting
% to knietic feasible trajectoreis, which is a multi-modal porblem with lots of
% hetregenoreity in camera images (number of cameras, camera positions),
% vehicle dynmaics, and so on. In terms of environement, vehicle embodiment and
% regulations. For envrionment, the driving can be on-road, off-road, or
% highway, urban, parking or geographically distributed like China, India,
% Japan or US. For vehicle dynamics and embodiment, it can be passenger
% vehicle, trucks, and off-road vechile (called UAV or what). For regulations,
% different regions have their own traffic rules, like Japan driving on left
% and China drives on right. This heterogeneity can not be handled by one model
% even with all domain data pooled. MoE can provide the solution. But classical
% MoE also needs pooled data to train an initialized sparse MoE architecture
% from scracth. This is not feasible in most use cases domain data separeted
% with privacy concerns as metioned in the previous point.
% -----------------------------------------------------------------------------

\begin{figure}[t]
\centering
\includegraphics[width=\columnwidth]{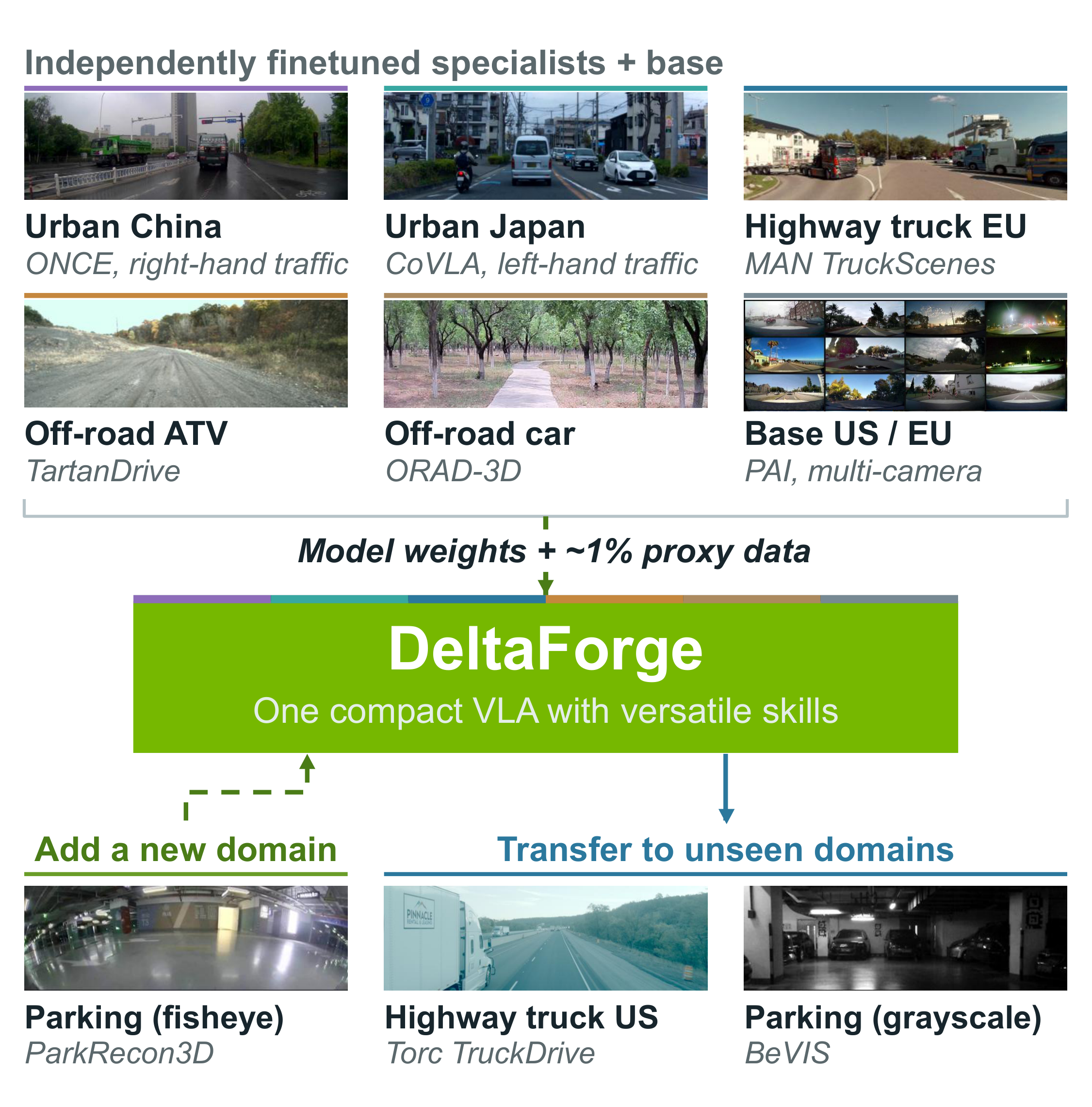}
\caption{\method{} consolidates independently finetuned per-domain driving VLAs
into one model from their weights and a small proxy set. A joint SVD factorizes the
per-domain deltas into shared skill bases and low-rank cores that clip-level
blind routers compose at inference.}
\label{fig:teaser}
\end{figure}

% ---- P4: Challenge C3 — efficiency ------------------------------------------
\textbf{Efficiency.}
\emph{MoErging} methods \cite{yadav2025moerging,ostapenko2024arrow} combine
independently finetuned experts into a single mixture-of-experts post hoc. Yet they route large dense blocks, so storage grows
almost linearly with the number of domains and top-$k$ mixing becomes barely
affordable. Classical sparse MoE is efficient, but its token- and layer-level
routers \cite{shazeer2017outrageously,fedus2022switch} suit language decoding,
whereas the waypoints of one trajectory share a single scene and vehicle, so
fine-grained switching adds complexity without benefit. A few post-hoc
parameter-efficient methods \cite{vasani2025phlora,kaushik2025eigenlorax} do
not address new-domain opt-in, and none target VLA models.
% --- original draft ----------------------------------------------------------
% efficiency: MoErging-style (cite) methods can provide solution to the
% privacy-preserving and data-availability MoE methods. While, they are
% typically very dense with part of parameters shared but most of parameters
% routed, which is not efficient for both fine-tuning and inference espicially
% barely possible for top-k routing. while top-K is important for making mixing
% different knowledge trained in each domain model barely possible. Classical
% MoE is sparse in terms of higher inference efficiency and mixing experts well
% but uses token-level or layer-level router that are normally seen for LLMs
% (citations). In AV scenarios, many waypoints/tokens in one trajectory
% prediction usually obey the same envrionment perception/reasoning and dynamic
% capabilities, without the needs to have token or layer-level routing which
% causes extra complexity to the model. This cast severe challenges for a
% efficent MoErging-style methods. Some of works use post-hoc
% parameter-efficent expert merging method. While their method is also not
% emphasized for new domain opt-in, an importatnt feature mention bleow point.
% Also none of them were applied on VLA model as metioned before.
% -----------------------------------------------------------------------------

% ---- P5: Challenge C4 — flexibility / new-domain opt-in ---------------------
\textbf{Flexibility.}
To be feasible, a deployed foundation model must accept new domains continually. Also, as partners keep contributing
weights, their knowledge must be injected without exponentially expanding dense parameters or
retraining the whole system each time. However, finetuning the
global model on each newcomer needs its dataset and risks catastrophic
forgetting, weight merging \cite{wortsman2022soups} loses effectiveness as
heterogeneous domains accumulate, and existing MoErging methods either forbid
expert opt-in or require end-to-end retraining. 
% --- original draft ----------------------------------------------------------
% flexibility: In the realistic appplication use cases, the global foundation
% model needs high expandibility. When a new client partnet domain decieds to
% opt in the global foundation model, it is challenging to flexibly and
% efficiently inject the new domain knowledge. Finetuning requires domain
% datasets. Also, theere might be catastrophical forgetting problem when a new
% finetune the base model. Causing the model forgets the pretrained base
% capability. Merging weight such as model soup can be a solution, while it may
% not be effective when there are manhy heterogeneous driving domains.
% MoErging-style methods could be ideal, while either they are not avaliable
% for new expert optin, or is not efficient requiring end-to-end finetuning
% with large dataset (cite). this is a challenge of continual learning When new
% client partner continuously agrees to share their model weights to the
% ceneralized model, how do we eifficiently opt-in model specialezd knowledge
% without introudcing linearly exploing dense paraemters and end-to-end
% finetunig every time.
% -----------------------------------------------------------------------------

% ---- P6: Research question — compositional generalization -------------------
Beyond these four challenges, consolidating domain models raises a question
of its own: \emph{do domain skills compose?} As experts accumulate, does the global
model gain capability even on unseen domains? For example, can a highway
trucking expert and an off-road passenger-vehicle expert be mixed into an
off-road trucking capability for construction scenarios? Such compositional
generalization has barely been studied outside single-modality language
tasks, and it is harder still for VLA driving models.
% --- original draft ----------------------------------------------------------
% generalization: Driving scenarios can be dynamic, ego vehicle can drive to
% different envrionmetns and have differnt dynamic features under differnt
% environemtns (higway, urban, offroad). As domain experts growing in the
% foundation model, does it increase the global model's multi-domain
% capabilities and generalization even on unseen domain tasks?  For exmple, if
% we have a highway truck domain model and a off-road passenger vehicle model,
% can we mix their capabilities to have unseen offroad trucking capability for
% construction scenarios? Very limited previous research focus on this novelty,
% or most of them focus on single modality task without much heterogeneity like
% LLM language tasks. While, this is even more challenging for VLA driving
% models.
% -----------------------------------------------------------------------------

% ---- P7: Our approach — DeltaForge -----------------------------------------
To tackle these challenges, we propose \method{}~(Fig.~\ref{fig:teaser}), a
driving MoE VLA built by post-hoc composition of independently finetuned domain
models, built on the Alpamayo VLA \cite{nvidia2025alpamayo}. Its core,
\emph{Post-hoc Expert Factorization} (PEF), exploits the low-rank nature of
finetuning \cite{hu2022lora,aghajanyan2021intrinsic} to jointly factorize the
domain deltas into a frozen \emph{shared skill basis} and per-domain
\emph{low-rank cores}, so routing moves tiny cores rather than dense blocks.
Two cooperating clip-level routers, one at the input for VLM reasoning experts
and one intermediate for flow-matching dynamics experts, mix cores without any
domain labels. A new domain opts in by
projecting its delta onto the frozen basis, with no end-to-end retraining.
% --- original draft ----------------------------------------------------------
% To tackle the above challenges, We propose a generliazed foundation driving
% VLA MoE model AlpaMoE, based on open-source Alpamayo a VLM plus flow matching
% style VLA model, following an intutive and simply efficient domain-fintuned
% expert merging, decoomposition, mixing-routing, new domain adding. The
% framwork is based on proposed generalizable post-hoc expert factorization
% (PEF) method, a parameter-efficient post-hoc epxert merging technique, to
% ensure the model inference and training efficiency for top-k routing and
% driving reasoning and dynamic experts composition (or mixing) for one global
% driivng model.
% In some literatures, it is found that SFT actually changes a small amount of
% parameters, Thus, PEFT methods is propsed such as lora (literature citation).
% Inspired by it, our method extracts common driving skills from domain
% finetuned model from the same pretrained base model, and keep specialized
% skill compostiion coeffcient as domain low-rank core experts. Then we frozen
% shared common skill bases and route low-rank core experts for efficient and
% flexible trianing and inference. This also enables flexible combination of
% vlm reasoning and flow matching diffusion dynamic driving skills with
% skill-mixing router.
% -----------------------------------------------------------------------------

% ---- P8: Contributions -------------------------------------------------------
Our contributions are threefold:
\begin{itemize}
  \item We propose \emph{\method{}}, to our knowledge the first post-hoc
        mixture-of-experts VLA for driving: it consolidates independently
        finetuned per-domain models into one multi-domain VLA from their
        weights and a $1\%$ proxy set, and later lets new domains opt in
        cheaply without retraining or sharing data.
  \item We introduce \emph{Post-hoc Expert Factorization} (PEF), which uses a
        joint SVD to split the per-domain finetuning deltas into a frozen
        shared skill basis and low-rank cores, together with a \emph{two-level
        skill-mixing router} that composes VLM reasoning and flow-matching
        dynamics experts at the clip level, so routing moves tiny cores rather
        than dense blocks.
  \item We evaluate on six domains that span urban, off-road, and highway
        driving on passenger cars, trucks, and ATVs across several regions.
        \emph{\method{}} approaches the per-domain experts at close to
        single-model cost, surpasses every merging, routing, and data-pooling
        baseline, and shows compositional generalization to unseen domains.
\end{itemize}

%% file: sections/02_related_work.tex
\section{Related Work}
\label{sec:related}

% Target length: <= half a page. Gap statements only (no DeltaForge pitch).
% All cite keys verified 2026-08-31 against literature PDFs or arXiv;
% do not add citations without verifying.

\subsection{Vision-Language-Action Models for Driving}
\label{sec:related:vla}
Driving VLAs couple a vision--language backbone with an action head for
end-to-end planning \cite{nvidia2025alpamayo}. MoE variants specialize experts
within a single model and a single training run, routing camera views,
behaviors, or scene-adaptive experts \cite{yang2025drivemoe,wan2025geminus}.
All of these train their experts jointly on pooled multi-domain data.
Consolidating independently finetuned driving models, or adding a domain
without retraining, is not addressed.

\subsection{MoE, Model Merging, and MoErging}
\label{sec:related:composition}
Composition during training requires pooled data: sparse MoE routes token-
or layer-level experts \cite{shazeer2017outrageously,fedus2022switch}, and
branch-train variants parallelize expert training but still merge or
re-finetune centrally \cite{sukhbaatar2024btx}. Post-hoc alternatives avoid
pooled data: weight averaging and task arithmetic
\cite{wortsman2022soups,ilharco2023task} are training-free but lose accuracy
as heterogeneous domains accumulate, while MoErging methods
\cite{yadav2025moerging,ostapenko2024arrow} route frozen experts at the cost
of storing dense per-expert parameters, which makes top-$k$ mixing expensive.
Closest in spirit, PHLoRA \cite{vasani2025phlora} extracts low-rank adapters
from full-finetuning deltas one checkpoint at a time, and EigenLoRAx
\cite{kaushik2025eigenlorax} recycles existing LoRA adapters into a shared
principal subspace. Both build on the low-rank view of fine-tuning~\cite{hu2022lora,aghajanyan2021intrinsic}, treating adapter directions as reusable structure across models. A jointly factorized shared basis
with routed per-domain cores, applied to the two components of a VLA,
remains unexplored.

\begin{figure*}[t]
\centering
\includegraphics[width=\textwidth]{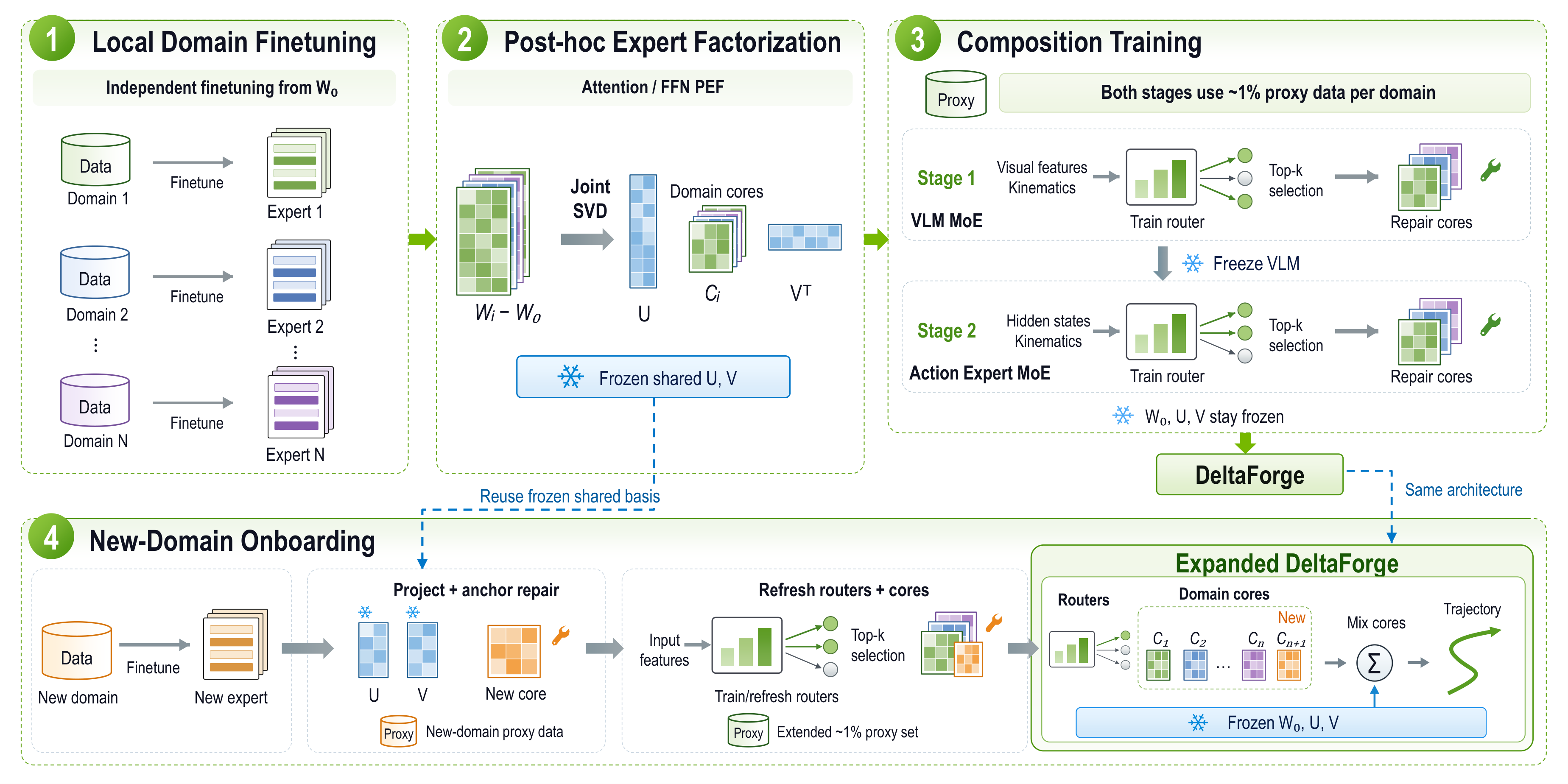}
\caption{\method{} system workflow. Independently finetuned per-domain VLAs are
jointly factorized by PEF into a frozen shared skill basis and per-domain
low-rank cores. Two cooperating clip-level blind routers then select and mix
the cores at inference, one at the input for VLM reasoning experts and one
intermediate for flow-matching dynamics experts. Composition is trained on a
small proxy set, and a new domain opts in by projecting its delta onto the
frozen basis.}
\label{fig:overview}
\end{figure*}

\subsection{Federated and Multi-Domain Finetuning}
\label{sec:related:fl}
% Federated finetuning exchanges weights instead of data
% \cite{mcmahan2017communication,ye2024openfedllm}, increasingly with LoRA
% adapters and heterogeneity-aware aggregation \cite{wang2024flora}, and has been
% applied to LLM-based driving \cite{xiang2025flad}.
Federated finetuning exchanges weights instead of data \cite{mcmahan2017communication,ye2024openfedllm}. Recent work incorporates LoRA adapters and heterogeneity-aware aggregation \cite{wang2024flora}, and the paradigm has been applied to LLM-based driving \cite{xiang2025flad}.
It still requires synchronized training
rounds across all clients, and the averaged global model loses accuracy under
strong domain heterogeneity. Closest to us, FlexOlmo \cite{shi2025flexolmo}
combines independently trained domain experts into an MoE without sharing data,
but routes over dense experts at the token level; it never \emph{factorizes}
them into the shared basis and compact routed cores that make our composition
efficient.

% --- original draft (first skeleton) -----------------------------------------
% 4 subsections: Driving VLA models / MoErging / Mixture-of-experts and model
% composition \todo{BTX, merging (soup/TIES/DARE), per-domain adapters.} /
% continual learning. Merged into 3 on 2026-08-31 (user decision).
% -----------------------------------------------------------------------------

%% file: sections/03_method.tex
\section{DeltaForge}
\label{sec:method}

% =============================================================================
% III-A  Problem Formulation and Overview        (+ Fig. 1 method overview)
% III-B  Post-hoc Expert Factorization (PEF)     (bold run-in paragraphs)
% III-C  Skill-Mixing Routing                    (bold run-in paragraphs)
% III-D  Proxy-Based Composition Training        (bold run-in paragraphs)
% III-E  Flexible New-Domain Onboarding
% (Parameter/efficiency accounting lives in IV-C Efficiency Analysis.)
% Original draft notes are kept verbatim in comments below each slot.
% =============================================================================

\subsection{Problem Formulation and Overview}
\label{sec:method:setup}
We consider the base weights $W_0$ and $N$ heterogeneous
domains (different environments and vehicle embodiments). Each domain $i$
holds a private dataset $\mathcal{D}_i$ and performs full-parameter finetuning on it,
\begin{equation}
  W_i \;=\; \operatorname*{arg\,min}_{W \,\gets\, W_0}\;
  \mathbb{E}_{(x,y)\sim\mathcal{D}_i}\,\mathcal{L}(W; x, y),
  \label{eq:domain_ft}
\end{equation}
where $W \!\gets\! W_0$ denotes optimization initialized at the shared base.
For consolidation, only the finetuned weights $\{W_i\}_{i=1}^{N}$ and a small
proxy subset of each training split are available: the bulk of
$\mathcal{D}_i$ never leaves its domain. We seek a composition operator
$\mathcal{C}$ that turns the domain models into one MoE with
\emph{multi-domain capability}:
\begin{equation}
  W_{\mathrm{MoE}} \;=\; \mathcal{C}\big(W_0,\, \{W_i\}_{i=1}^{N}\big).
  \label{eq:moe_goal}
\end{equation}

Concretely, the goal is threefold: (i) \emph{capability match}: on each
domain, $W_{\mathrm{MoE}}$ matches the corresponding expert $W_i$;
(ii) \emph{blind efficient routing}: given a new input without any domain
label, the MoE identifies it and efficiently routes and combines the relevant
reasoning and dynamics skills; (iii) \emph{flexible growth}: new domains opt
in cheaply, and the
composed model generalizes beyond the seen domains.

We instantiate \method{} on Alpamayo \cite{nvidia2025alpamayo}, an
open-source driving VLA
with two components: a VLM and a flow-matching action expert, finetuned in
two stages, so each domain contributes a reasoning expert and a dynamics
expert. \method{} generalizes to any two-component VLA of this
reasoning-plus-action form. Figure~\ref{fig:overview} gives the overall system
workflow.

% --- original draft ----------------------------------------------------------
% setup: heterogeneous domains (environments, embodiment). Each domain full
% finetune from the same pretrained model. Then composite domain models into
% MoE which has multi-domain capabilities.
% Alpamayo architecture: two-stage finetuning vlm and diffusion.
% -----------------------------------------------------------------------------

\subsection{Post-hoc Expert Factorization (PEF)}
\label{sec:method:pef}

Finetuning updates are effectively low-rank
\cite{hu2022lora,aghajanyan2021intrinsic}, so the $N$ experts are highly
redundant and routing full expert
blocks is needlessly dense. Basically, low-rank structure means the experts share significant overlap, so routing full blocks is wasteful. PEF exploits this structure to
keep the MoE sparse and cheap: it extracts the skills shared across domains
into a common frozen basis, keeps one small core expert per domain that
encodes its skill combination, and routes only the attention and FFN cores.

\begin{figure*}[t]
\centering
\includegraphics[width=0.95\textwidth]{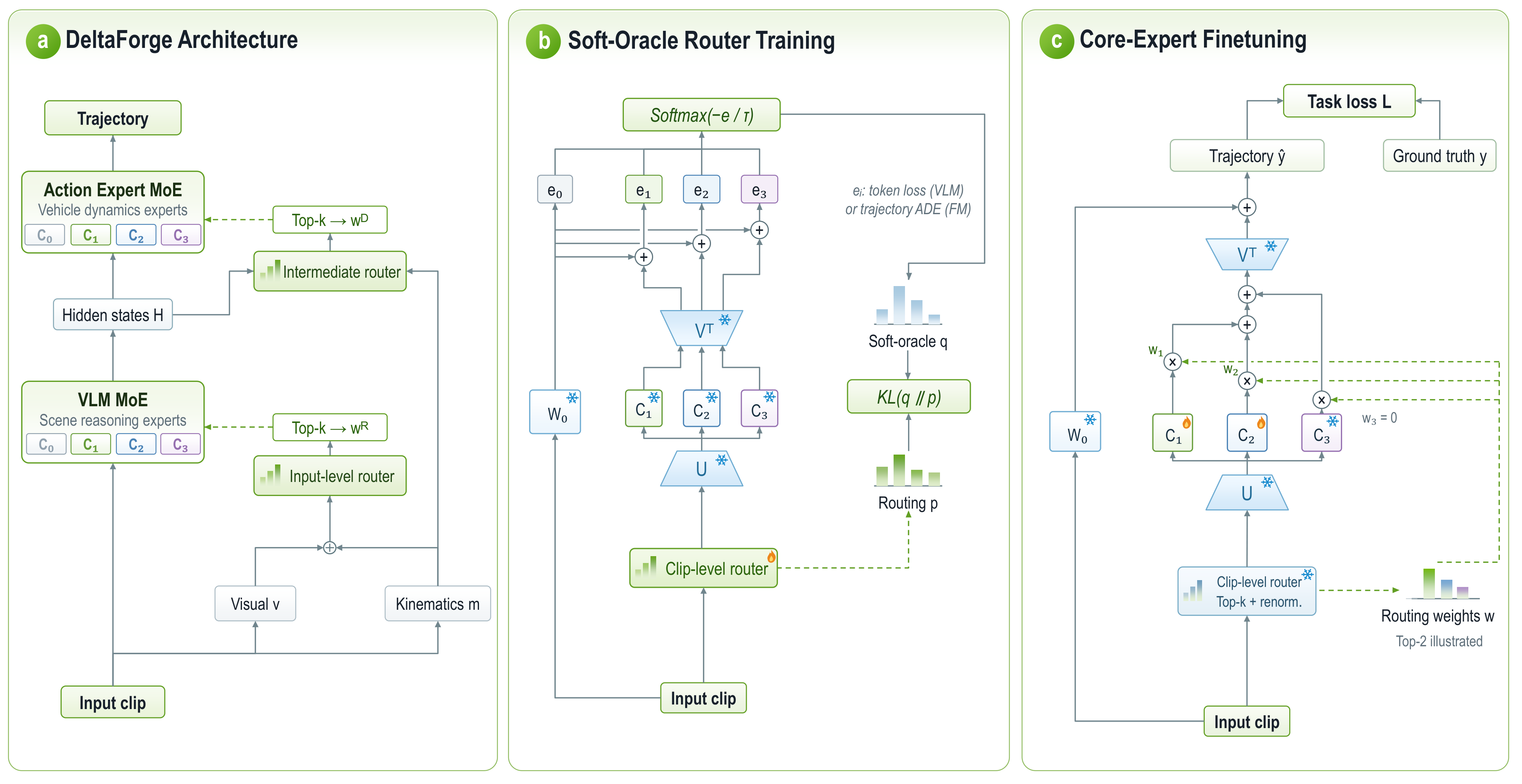}
\caption{\method{} architecture and composition training. (a) Two-level
clip-level routing of low-rank cores at inference. (b) Soft-oracle router
training with a KL objective. (c) Core-expert finetuning against the task
loss.}
\label{fig:arch}
\end{figure*}

\textbf{Shared skill basis via joint SVD.}
For every factorized weight matrix
$W \in \mathbb{R}^{d_{\mathrm{out}} \times d_{\mathrm{in}}}$ (acting as
$x \mapsto Wx$), each domain's update is first summarized by its task vector
\cite{ilharco2023task}
\begin{equation}
  \Delta_i \;=\; W_i - W_0, \qquad i = 1, \ldots, N .
  \label{eq:taskvec}
\end{equation}
We then extract a joint rank-$r$ basis from all $N$ task vectors,
\begin{equation}
\begin{aligned}
  U &= \operatorname{SVD}_r\!\big([\Delta_1, \ldots, \Delta_N]\big),\\
  V &= \operatorname{SVD}_r\!\big([\Delta_1^{\top}, \ldots, \Delta_N^{\top}]\big),
\end{aligned}
  \label{eq:jointsvd}
\end{equation}
where $[\cdot]$ concatenates along columns and $\operatorname{SVD}_r$
returns the top-$r$ left singular vectors of its argument. By the
Eckart--Young theorem \cite{eckart1936approximation}, truncated SVD is the
optimal low-rank
approximation, so $U \in \mathbb{R}^{d_{\mathrm{out}} \times r}$ and
$V \in \mathbb{R}^{d_{\mathrm{in}} \times r}$ span the best shared column and
row spaces of all domain updates at a low rank
$r \ll \min(d_{\mathrm{out}}, d_{\mathrm{in}})$. We interpret the frozen pair $(U, V)$ as a \emph{shared skill basis} whose size
is fixed by $r$ and independent of $N$.
% --- original draft ----------------------------------------------------------
% Shared skill base by joint SVD: compute task vector on each expert
% ($w-w_0$), extract skills.
% -----------------------------------------------------------------------------

\textbf{Per-domain low-rank core experts.}
Each domain keeps only its coordinates in this basis, the \emph{core expert}
\begin{equation}
  C_i = U^{\top} \Delta_i V \in \mathbb{R}^{r \times r},
  \qquad
  \widehat{W}_i = W_0 + U C_i V^{\top},
\end{equation}
which encodes how domain $i$ combines the shared skills. The basis is frozen
and shared across all domains. Only the small cores are stored, routed, and
finetuned. The base model itself joins the expert pool for free as a zero
core, $C_0 = \mathbf{0}$ (so $\widehat{W}_0 = W_0$), giving the routers a
default base route.
% --- original draft ----------------------------------------------------------
% Per-domain low-rank core experts: keep coefficients of skill combination as
% each low-rank core expert. Freezing and share base skill matrix, route the
% low-rank core experts.
% -----------------------------------------------------------------------------

\textbf{Factorization scope.}
PEF is applied to the attention and FFN matrices of the ViT, the LLM, and the
flow-matching action expert alike, where routing low-rank cores is highly
parameter-efficient. The remaining modules (token embeddings, LM head,
etc.)\ are merged into a single shared copy and not routed.
% --- original draft ----------------------------------------------------------
% Merge or route: PEF on both attention and FFNs in ViT, LLM (as very
% efficient after PEF), merge the rest of them (embed, lm head, and so on).
% -----------------------------------------------------------------------------

\subsection{Skill-Mixing Routing}
\label{sec:method:routing}

\method{} composes the two expert families of
Sec.~\ref{sec:method:setup}, VLM reasoning experts and flow-matching dynamics
experts, with two cooperating routers~(Fig.~\ref{fig:arch}a) so that reasoning and
dynamics skills can be mixed across domains.
% --- original draft ----------------------------------------------------------
% Two-stage MoE composition: stage-1 VLM and stage-2 flow matching.
% -----------------------------------------------------------------------------

\textbf{Clip-level blind routing.}
The router computes one expert mixture $w$ per input clip, and every factorized layer $\ell$ mounts the same
mixture for the entire trajectory:
\begin{equation}
  \bar{W}^{(\ell)} \;=\; W_0^{(\ell)} +
  U^{(\ell)} \Big(\textstyle\sum_{i=1}^{N} w_i\, C_i^{(\ell)}\Big)
  V^{(\ell)\top} .
  \label{eq:cliprouting}
\end{equation}
This departs from the per-token and per-layer routers of LLM MoEs by design:
all waypoints of one trajectory obey the same scene and vehicle dynamics, so
finer-grained switching adds complexity without benefit.
% --- original draft ----------------------------------------------------------
% Top-k routing, clip-level. Router is clip-level, for each input, a set of
% experts are chosen to generate all tokens for this trajectory. This is
% different from LLM's per-token or per-layer router, because trajetores are
% for clip-level.
% -----------------------------------------------------------------------------

\textbf{Input-level router for VLM reasoning experts.}
The VLM router is blind: it never sees a domain label. Its two inputs are
\begin{equation}
  v \;=\; \tfrac{1}{|\mathcal{T}|} \textstyle\sum_{t \in \mathcal{T}}
  \mathrm{ViT}_0(x)_t,
  \qquad
  m \;=\; \big[\phi(s),\, \phi(a),\, \phi(\kappa)\big],
  \label{eq:router_feats}
\end{equation}
where $\mathrm{ViT}_0(x)_t$ are the visual token features of the frozen ViT
of the base VLM $W_0$ ($\mathcal{T}$ the visual tokens), $s, a, \kappa$ are
the speed,
acceleration, and curvature sequences of the ego history, and $\phi$ returns
their z-scored summary statistics (mean, standard deviation, extrema, final
value). A small MLP then yields
$p^{\mathrm{R}} = \operatorname{softmax}\!\big(f_{\theta}([v;\, m])\big)$.
Together, $v$ and $m$ identify the scene environment and the vehicle
embodiment.
% --- original draft ----------------------------------------------------------
% Two routers (one at input-level for VLM, and one intermediate for diffusion)
% enable flexible combination of reasoning (VLM) and dynamic (diffusion)
% experts. Router never sees a domain label (blind input-based router). Router
% input is visual features from base ViT, and dynmic features (speed,
% acceleration, stats; normalized) for scene environment and vehicle
% embodiment identification.
% -----------------------------------------------------------------------------

\textbf{Intermediate router for trajectory dynamics experts.}
The dynamics router reads $h$, the mean-pooled last-layer hidden states of
the VLM,
together with the same kinematics $m$, and selects dynamics experts via
$p^{\mathrm{D}} = \operatorname{softmax}\!\big(f_{\phi}([h;\, m])\big)$.
Decoupling the two routers lets one domain's scene reasoning pair freely with
another domain's vehicle dynamics.
% --- original draft ----------------------------------------------------------
% Intermediate router uses output of VLM hidden state and dynamic features as
% input, routing diffusion dynamic experts.
% -----------------------------------------------------------------------------

\textbf{Top-$k$ soft mixing.}
Each router distribution $p$ is truncated to its top-$k$ experts
$\mathcal{K}$ and renormalized to weights $w$ over that set, with $w_i{=}0$
outside $\mathcal{K}$.
Since Eq.~(\ref{eq:cliprouting}) is linear in the cores, mixing happens once
in parameter space at zero extra inference cost, and $w$ stays in the
computation graph, keeping routing end-to-end trainable. Top-$1$, top-$2$,
and top-$p$ are instances of the same mechanism.

\subsection{Proxy-Based Composition Training}
\label{sec:method:training}

All post-hoc training of \method{} uses a very lightweight proxy set
$\mathcal{P}_i \subset \mathcal{D}_i$ per domain, roughly $1\%$ of the
domain's training data, and it serves only two purposes: router training and
core-expert repair.
% --- original draft ----------------------------------------------------------
% All the PEF training are based on a small proxy train set (200 per domain
% compared to 20K+ total training samples).
% -----------------------------------------------------------------------------

\textbf{Soft-oracle router training.}
Router supervision is generated by the experts themselves~(Fig.~\ref{fig:arch}b). No domain label is
ever used. Every candidate expert scores each proxy sample,
\begin{equation}
  e_i(x) \;=\;
  \begin{cases}
    \mathcal{L}\big(\widehat{W}_i;\, x, y\big), & \text{VLM experts,}\\[2pt]
    \mathrm{ADE}\big(\hat{y}_i(x),\, y\big), & \text{flow-matching experts,}
  \end{cases}
  \label{eq:expertscore}
\end{equation}
the teacher-forced token loss for reasoning experts, and the error of the
generated trajectory $\hat{y}_i(x)$ for dynamics experts, whose
flow-matching training loss correlates poorly with trajectory quality. The
sweep is cheap thanks to PEF: all experts share frozen skill basis and
differ only by their $r \times r$ cores. The scores define a loss-minimizing
soft oracle label with temperature $\tau$,
\begin{equation}
  q_i(x) \;=\;
  \frac{\exp\!\big(-e_i(x)/\tau\big)}
       {\sum_{j=1}^{N} \exp\!\big(-e_j(x)/\tau\big)},
  \label{eq:softoracle}
\end{equation}
and the router is trained to match this soft-oracle distribution by minimizing
the KL divergence
\begin{equation}
  \mathbb{E}_{x \sim \cup_i \mathcal{P}_i}\,
  \mathrm{KL}\big(q(x)\,\|\,p(x)\big)
  \;=\;
  \mathbb{E}_{x}\!\sum_{i=1}^{N} q_i(x)\,
  \log\tfrac{q_i(x)}{p_i(x)},
  \label{eq:routerkl}
\end{equation}
where $p(x)$ is the corresponding router output ($p^{\mathrm{R}}$ or
$p^{\mathrm{D}}$). The calibrated target natively supports top-$k$
selection.
% --- original draft ----------------------------------------------------------
% Router training uses soft oracle labels by teacher-forcing method. each
% sample does forward pass on each experts and records loss-minimizing
% experts. The labels are constructed using softmax with temperature, then the
% soft labels are used for router training with KL divergence loss. The number
% of expert cores that are jointly finetuned is according to the top-k
% strategy.
% -----------------------------------------------------------------------------

\textbf{Core-expert finetuning.}
Core-expert finetuning on the proxy set can potentially compensate for the rank-$r$ SVD truncation
loss~(Fig.~\ref{fig:arch}c). Once the router is trained, we freeze it and optimize the cores
selected by its top-$k$ policy. The shared skill basis $(U, V)$ also stays
frozen, so among the model weights only the core experts $\{C_i\}$ are
optimized by minimizing the task loss $\mathcal{L}$ over the cores on the
proxy $\cup_i \mathcal{P}_i$, mounting the routed core mixture
$\bar{W}(x; \{C_i\})$ of Eq.~(\ref{eq:cliprouting}). Gradients reach only the top-$k$ selected
cores, and learning under the top-$k$ mixture lets these cores
co-adapt, improving expert mixing at inference.
% --- original draft (as edited on Overleaf) -----------------------------------
% We keep the shared skill bases U/V frozen, and only fine-tune core experts
% for recovery repair after SVD (it actually performs as regularization).
% -----------------------------------------------------------------------------

\textbf{Two-stage composition training.}
We first train the VLM MoE (the input-level router and the VLM cores), then
freeze it and train the flow-matching MoE with the intermediate router. The
routers thereby learn by themselves how to combine reasoning and dynamics
experts across domains.
% --- original draft ----------------------------------------------------------
% Two-stage training for vlm and diffusion. train moe vlm first, then frozen
% moe vlm and train diffusion moe. So that the routers learned themselves how
% to combine reasoning and dynamic skills/experts.
% -----------------------------------------------------------------------------

\subsection{Flexible New-Domain Onboarding}
\label{sec:method:onboarding}
A new domain $N{+}1$ opts in against the \emph{frozen} skill basis. The basis
$(U, V)$ stays frozen and no domain is retrained end-to-end. First, its task vector is
projected onto the frozen basis to extract a new core expert,
\begin{equation}
  \Delta_{N+1} \;=\; W_{N+1} - W_0,
  \qquad
  C_{N+1} \;=\; U^{\top} \Delta_{N+1} V .
  \label{eq:onboard}
\end{equation}
Second, the new core is anchor-repaired on its own proxy set
$\mathcal{P}_{N+1}$ by core repair with routing locked to expert
$N{+}1$, compensating for the lossy projection onto a basis built without
this domain. Third, soft-oracle labels are regenerated
(Eq.~\ref{eq:softoracle}) and the routers and all cores are jointly
refreshed by re-running router training (Eq.~\ref{eq:routerkl}) and core repair with
$N \!\gets\! N{+}1$ on the extended proxy set
$\cup_{i=1}^{N+1} \mathcal{P}_i$, which adapts the expert mixtures to the
new domain while preserving performance on the old ones.

% Nothing else changes---in contrast to rebuilding the
% entire MoE (re-running the joint SVD and retraining all components) whenever a
% domain is added. Moreover, before any adaptation, the frozen MoE can already
% serve an unseen domain zero-shot by mixing existing experts
% (leave-one-domain-out), which we evaluate in Sec.~\ref{sec:exp:onboarding}.
% --- original draft ----------------------------------------------------------
% Flexible new domain on boarding: based on the current frozen skill base U/V,
% do SVD on new fine-tuned domain model and keep core expert. Opt in 200
% domain proxy dataset. Frozen skill base and anchored repair core with proxy,
% then use whole proxy dataset to generate soft oracle labels and jointly
% finetune the router and core experts.
% Zero-shot capability of our MoE (leave one domain out).
% -----------------------------------------------------------------------------

%% file: sections/04_experiments.tex
\section{Experiments}
\label{sec:experiments}

% =============================================================================
% IV-A  Experimental Setup      (base model; domains Table I; protocol)
% IV-B  Main Results: Multi-Domain Performance   (baselines + Table II)
% IV-C  Efficiency Analysis                      (Table III)
% IV-D  New-Domain Onboarding and Zero-Shot Generalization
% IV-E  Ablations and Analysis
% =============================================================================

\subsection{Experimental Setup}
\label{sec:exp:setup}

\textbf{Base model and training.}
We build on Alpamayo 1.5 (10B) \cite{nvidia2025alpamayo}, an 8B Cosmos-Reason VLM backbone with a 2B flow-matching action expert. In its default trajectory mode, the model takes a front camera image and a 16-step ego history at 10Hz and predicts a 64-step future trajectory. Each domain expert is trained in two stages on eight
H100 GPUs: stage~1 fully finetunes the VLM with next-token cross-entropy on
trajectory tokens (AdamW, lr $10^{-5}$, batch 32), and stage~2 freezes it and
trains the action expert with the flow-matching objective (lr $10^{-4}$, batch
8), each up to 10 epochs. For composition, the PEF cores have rank $r{=}256$,
the soft-oracle temperature is $\tau{=}0.3$, and the routers and cores are
trained for 200 and 500 steps on the proxy set.
% --- original draft ----------------------------------------------------------
% Base model: Alpamayo 1.5 end-to-end driving reasoning model, Qwen3-VL
% backbone + flow matching/diffusion. Two-stage training details:
% hyperparameters.
% -----------------------------------------------------------------------------

\textbf{Domains datasets.}
Table~\ref{tab:domains} lists the six domains: five finetuned experts and the
base model's native domain, served by the zero-core route. They are
heterogeneous in environment, region, vehicle, and sensor rig, and each
finetuned domain uses one front camera while the base uses four. Three further
datasets are held out for Sec.~\ref{sec:exp:onboarding}: ParkRecon3D for
new-domain onboarding, and Torc TruckDrive and BeVIS for zero-shot
generalization.

\begin{table}[tb]
\centering
\caption{Domains and datasets. Train/eval are windowed samples after
scene-level splitting. The lower block is used only for onboarding and
zero-shot evaluation.}
\label{tab:domains}
\scriptsize
\setlength{\tabcolsep}{3.5pt}
\begin{tabular}{@{}lllrr@{}}
\toprule
Domain & Dataset & Scenario & Train & Eval \\
\midrule
Urban China      & ONCE~\cite{mao2021once}                     & urban, right-hand      & 24,885 & 2,769 \\
Urban Japan      & CoVLA~\cite{arai2024covla}                  & urban, left-hand       & 26,987 & 2,893 \\
Off-road ATV     & TartanDrive~\cite{sivaprakasam2024tartandrive} & forest trails, ATV       & 17,470 & 1,484 \\
Off-road car     & ORAD-3D~\cite{min2025orad3d}                & China, off-road, car   & 34,801 & 4,897 \\
Highway truck EU & TruckScenes~\cite{fent2024truckscenes}      & EU highway, truck      & 33,305 & 3,172 \\
Base             & PhysicalAI-AV~\cite{nvidia2025pai}          & US/EU, four cameras           & 19,545 & 3,000 \\
\midrule
Highway truck US    & Torc TruckDrive~\cite{ghilotti2026truckdrive} & US highway, truck   & 23,593 & 3,031 \\
Underground parking & ParkRecon3D~\cite{wei2026parkgaussian}       & garage, fisheye     & 5,753  & 745 \\
Indoor parking      & BeVIS~\cite{shao2023bevis}                   & garage, grayscale   & 4,485  & 297 \\
\bottomrule
\end{tabular}
\end{table}
% --- original draft ----------------------------------------------------------
% Domains: datasets(default trajectory end-to-end training), number of
% cameras, split, description, and others.
% -----------------------------------------------------------------------------

\textbf{Protocol and metrics.}
All post-hoc training uses a proxy set of 200 clips per domain (about 1\% of
the training data) drawn from the training split. The eval splits are
never touched. We report the minimum average displacement
error over six sampled trajectories at the 5\,s horizon, in meters
(\minade{}@5\,s). Each router selects among $N{+}1$ routes, the $N$ domain experts
plus the default base route (Sec.~\ref{sec:method:pef}).
% --- original draft ----------------------------------------------------------
% Protocol: metrics (min_ADE at 5 s, best-of-6), proxy set size, eval-split
% isolation.
% -----------------------------------------------------------------------------

% --- original draft (baselines note) ----------------------------------------
% Zero-shot base; per-domain full finetune (upper bound); merging family
% (model soup, TIES, DARE); other SOTA post-hoc composition methods.
% -----------------------------------------------------------------------------

\begin{table*}[!tbp]
\centering
\caption{End-to-end \minade{}@5\,s (m, best-of-6, lower is better) on the six
domains and their mean. PAI is the base domain. Data is the training data a
method needs beyond the finetuned weights. \textbf{Bold} and
\underline{underline} mark the best and second-best weight-based method per
column.}
\label{tab:main}
\footnotesize
\setlength{\tabcolsep}{3pt}
\begin{tabular}{@{}lccccccccccc@{}}
\toprule
\multirow{2}{*}{Method} & \multirow{2}{*}{Venue} & \multicolumn{7}{c}{\minade{}@5\,s (m) $\downarrow$} & \multirow{2}{*}{Data} & \multicolumn{2}{c}{Params (B) $\downarrow$} \\
\cmidrule(lr){3-9} \cmidrule(l){11-12}
 & & \shortstack{PAI\\{\scriptsize base}} & \shortstack{ONCE\\{\scriptsize China}} & \shortstack{CoVLA\\{\scriptsize Japan}} & \shortstack{Tartan\\{\scriptsize ATV}} & \shortstack{ORAD\\{\scriptsize off-road}} & \shortstack{TruckScenes\\{\scriptsize highway truck}} & \shortstack{Mean\\{\scriptsize~}} & & \shortstack{Active\\{\scriptsize~}} & \shortstack{Total\\{\scriptsize~}} \\
\midrule
Zero-shot base (lower bound)     & -- & 0.9568 & 2.1763 & 4.4319 & 2.5019 & 1.0800 & 6.6409 & 2.9646 & none & 11.08 & 11.08 \\
Per-domain experts (upper bound) & -- & 0.9568 & 1.0354 & 1.6828 & 1.2294 & 0.5141 & 0.4601 & 0.9798 & own domain & 11.08 & 66.47 \\
\midrule
Centralized SFT (pooled data)    & -- & 0.8098 & 1.2461 & 1.7701 & 1.3092 & 0.9206 & 0.6217 & 1.1129 & all & 11.08 & 11.08 \\
Federated SFT (FedAdam~\cite{reddi2021fedadam}) & ICLR'21 & 0.8076 & 1.2237 & 1.7172 & 1.3457 & 0.6991 & 0.8585 & 1.1086 & all (local) & 11.08 & 11.08 \\
\midrule
Model soup \cite{wortsman2022soups}        & ICML'22 & 1.4376 & 1.4184 & 2.0058 & 1.3878 & 0.9029 & 2.6261 & 1.6298 & none & 11.08 & 11.08 \\
Task arithmetic \cite{ilharco2023task}     & ICLR'23 & 1.8665 & 1.5162 & 1.9275 & 1.3994 & 1.0440 & 2.6462 & 1.7333 & 1\% proxy & 11.08 & 11.08 \\
TIES \cite{yadav2023ties}                  & NeurIPS'23 & 1.9396 & 1.4604 & 2.0902 & 1.2396 & 0.9853 & 3.8695 & 1.9308 & 1\% proxy & 11.08 & 11.08 \\
DARE \cite{yu2024dare}                     & ICML'24 & 1.4542 & 1.4153 & 1.9283 & 1.4088 & 0.8618 & 2.5378 & 1.6010 & 1\% proxy & 11.08 & 11.08 \\
TSV-M \cite{gargiulo2025tsv}               & CVPR'25 & 1.4930 & 1.4398 & 2.1516 & 1.4122 & 0.7582 & 2.6390 & 1.6490 & 1\% proxy & 11.08 & 11.08 \\
KnOTS \cite{stoica2025knots}               & ICLR'25 & 2.0984 & 1.3874 & 1.9547 & 1.2497 & 0.7457 & 3.1853 & 1.7702 & 1\% proxy & 11.08 & 11.08 \\
EMR-Merging \cite{huang2024emr}            & NeurIPS'24 & \underline{0.9110} & 1.2632 & 1.9057 & 1.2181 & 0.6440 & 1.6974 & 1.2732 & 1\% proxy & 11.08 & 22.2 \\
Twin-Merging \cite{lu2024twin}             & NeurIPS'24 & 1.8385 & 1.3095 & 1.8514 & \underline{1.1764} & \underline{0.6321} & 1.2983 & 1.3510 & 1\% proxy & 11.12 & 11.30 \\
BTX \cite{sukhbaatar2024btx}               & arXiv'24 & 1.1892 & 1.2689 & \underline{1.7121} & 1.2049 & \textbf{0.5968} & 1.4668 & 1.2398 & 1\% proxy & 11.08 & 47.4 \\
\midrule
\textbf{\method{} top-1 routing}    & Ours & 0.9189 & \textbf{1.0185} & 1.7146 & 1.2053 & 0.6951 & \textbf{0.5417} & \textbf{1.0157} & 1\% proxy & 12.38 & 12.55 \\
\textbf{\method{} top-2 routing}    & Ours & \textbf{0.7569} & \underline{1.1806} & \textbf{1.7033} & \textbf{1.1727} & 0.6354 & \underline{0.6529} & \underline{1.0170} & 1\% proxy & 12.42 & 12.55 \\
\bottomrule
\end{tabular}
% Pooled row = final checkpoint (79.7k steps, no checkpoint selection). A
% mid-training checkpoint (24k steps) evaluated on the same subsets gives
% 0.8273/1.2423/1.6271/1.1754/0.5436/0.5271 (Mean-6 0.9905, Mean-5 1.0231);
% reporting it would mean selecting on the evaluation subsets. USER DECISION.
\end{table*}

\subsection{Main Results: Multi-Domain Performance}
\label{sec:exp:main}
\textbf{Baselines.}
Table~\ref{tab:main} compares four groups. The zero-shot base and per-domain
experts provide lower and upper bounds. Centralized and federated SFT
(FedAdam \cite{reddi2021fedadam}) use all training data, unlike the remaining
methods, which merge or route the same five checkpoints from weights and a
1\% proxy alone: static merging
\cite{wortsman2022soups,ilharco2023task,yadav2023ties,yu2024dare,gargiulo2025tsv,stoica2025knots}
and routed methods that keep the experts, EMR-Merging \cite{huang2024emr},
Twin-Merging \cite{lu2024twin}, and BTX \cite{sukhbaatar2024btx}. Merging
hyperparameters are selected on the proxy set.

\textbf{Results.}
\method{} is the most accurate weight-based method. Its mean \minade{}@5\,s is
1.0157, close to the 0.9798 of the six separate domain experts and below every
merging and routing baseline. Centralized and federated SFT do worse even though
they use all the training data, which shows how heterogeneous the domains are: one
shared model must trade them off, while routed experts do not. \method{} keeps the
base route, so it avoids the catastrophic forgetting most merging baselines suffer
on the base domain. Top-2 routing softly mixes two experts per clip, matching the top-1 mean while
cutting the base-domain error from $0.92$ to $0.76$ and adding robustness where
a single expert is weak.

% --- original draft ----------------------------------------------------------
% Comparison with baseline and other SOTA methods (one table).
% rop-1, top-2, top-P routing.
% -----------------------------------------------------------------------------

\subsection{Efficiency Analysis}
\label{sec:exp:efficiency}
\method{} is efficient in memory, compute, and training
(Tables~\ref{tab:params} and~\ref{tab:efficiency}).
Each domain adds a low-rank core of only 42M parameters, $0.3\%$ of the base.
Keeping separate experts or BTX FFN experts instead grows the model linearly,
so it soon becomes far larger than \method{}. At
inference, compute stays near single-model cost and barely grows with the
number of routed experts $k$: mixing low-rank cores keeps top-$k$ routing far
cheaper than dense MoEs. For training, we freeze the base and update only the cores and
routers, which is $1\%$ of the parameters, using the $1\%$ proxy dataset.

\begin{table}[tb]
\centering
\caption{Parameter breakdown of the deployed \method{} system ($N{=}5$).
The shared bases come in closed form from the joint SVD, so only the cores and
the two routers are trained, about $1\%$ of the total.}
\label{tab:params}
\footnotesize
\setlength{\tabcolsep}{4pt}
\begin{tabular}{@{}lrl@{}}
\toprule
Component & Params & Trained \\
\midrule
Base VLM + action expert          & 11{,}079\,M  & frozen \\
Shared bases $U,V$ (VLM + FM)      & 1{,}257\,M   & closed form \\
Embeddings, heads, norms          & +0           & merged \\
Routers ($\times 2$)              & 2.1\,M       & \checkmark \\
Cores + projections / domain      & 42.0\,M      & \checkmark \\
\midrule
Total ($N{=}5$)                   & 12{,}547\,M  &  \\
\quad of which trained            & 122.7\,M     & 1.0\% \\
\bottomrule
\end{tabular}
\end{table}

\begin{table*}[!tbp]
\centering
\caption{Deployment cost versus the number of domains $N$ and routed experts
$k$ (dense base 20.9 GFLOPs/token). \method{} stays near single-model cost as
both grow, unlike the linear scaling of separate experts and BTX.}
\label{tab:efficiency}
\footnotesize
\setlength{\tabcolsep}{6pt}
\begin{tabular}{@{}lrrrrrrr@{}}
\toprule
 & \multicolumn{3}{c}{Params (B) $\downarrow$} & \multicolumn{4}{c}{GFLOPs / token $\downarrow$} \\
\cmidrule(lr){2-4} \cmidrule(l){5-8}
Method & formula & $N{=}5$ & $N{=}10$ & formula & top-1 & top-2 & top-5 \\
\midrule
Per-domain experts & $11.08\,(N{+}1)$ & 66.5 & 121.9 & $20.9$ & 20.9 & -- & -- \\
BTX      & $11.08 + 7.26\,N$  & 47.4 & 83.7  & $6.35 + 14.52\,k$ & 20.9 & 35.4 & 79.0 \\
\textbf{\method{} (ours)} & $12.34 + 0.042\,N$ & 12.55 & 12.76 & $23.39 + 0.08\,k$ & 23.5 & 23.6 & 23.8 \\
\bottomrule
\end{tabular}
\end{table*}
% Sources: paper_final/results/EFFICIENCY_ANALYSIS.md (verified 2026-09-08);
% composition wall-clock from sacct (S1 J1 22 min, J2 23 min, J3 4h39;
% S2 K1 6 min, K2 1h33); ONCE stage-1 train_runtime 53,451 s (14.8 h) +
% stage-2 10-epoch run ~5 h.
% --- original draft ----------------------------------------------------------
% Efficiency analysis (one table).
% -----------------------------------------------------------------------------

\subsection{Zero-Shot Generalization and New-Domain Onboarding}
\label{sec:exp:onboarding}
We test \method{} on domains outside the training pool, using the
five-domain system of Table~\ref{tab:main}.

\textbf{Zero-shot skill transfer.}
When the pool already holds a related skill, an unseen domain benefits with no
training (Table~\ref{tab:generalize}, top). Torc TruckDrive is long-haul
highway trucking, which shares dynamics with MAN TruckScenes. 
% The blind router
% sends $90\%$ of its clips to the truck skill with top-1 routing and $83\%$ with
% top-2, and the error drops from 8.99 to 1.06 (top-1) or 1.00 (top-2), close to the domain's own
% finetuned expert (0.82) without any Torc TruckDrive data. A domain with no
% related skill in the pool shows no such gain.

The blind router sends $90\%$of its clips to the truck skill with top-1 routing and $83\%$ with top-2. The error drops from 8.99 to 1.06 (top-1) or 1.00 (top-2), approaching the domain's own finetuned expert (0.82), all without any Torc TruckDrive training data. A domain with no related skill in the pool shows no such gain.

\textbf{Onboarding.}
A new domain opts in against the frozen basis without touching the existing
experts or the basis (Sec.~\ref{sec:method:onboarding}). Onboarding ParkRecon3D
lowers its error from 0.75 to 0.65 and grows the system by only $42$M
parameters, while the six existing domains change by only $+1.5\%$ on average.
Onboarding also lifts a related domain that is still unseen
(Table~\ref{tab:generalize}, bottom): the never-seen BeVIS garages improve from
0.52 to 0.41, below BeVIS's own finetuned expert (0.42). A newly added parking
domain thus composes with the existing experts to serve a different, unseen
parking domain, answering the question raised in Sec.~\ref{sec:intro}.

\begin{table}[tb]
\centering
\caption{Generalization to unseen domains (\minade{}@5\,s, meters). Top: a
truck skill already in the pool (MAN TruckScenes) transfers zero-shot to the
unseen Torc TruckDrive. Bottom: onboarding the related ParkRecon3D parking
domain serves the never-seen BeVIS below its own finetuned expert, without
training on it.}
\label{tab:generalize}
\footnotesize
\setlength{\tabcolsep}{6pt}
\begin{tabular}{@{}lr@{}}
\toprule
\multicolumn{2}{@{}l}{\emph{Unseen:} Torc TruckDrive \quad \emph{in pool:} MAN TruckScenes} \\
\midrule
Model & \minade{} \\
\midrule
Base (zero-shot)      & 8.9862 \\
Own-domain finetune   & 0.8225 \\
\method{} top-1 routing      & 1.0627 \\
\method{} top-2 routing      & \textbf{0.9969} \\
\midrule
\multicolumn{2}{@{}l}{\emph{Unseen:} BeVIS \quad \emph{onboarded:} ParkRecon3D} \\
\midrule
Base (zero-shot)          & 0.5333 \\
Own-domain finetune       & 0.4238 \\
\method{} w/o ParkRecon3D & 0.5246 \\
\method{} w/ ParkRecon3D  & \textbf{0.4114} \\
\bottomrule
\end{tabular}
\end{table}

\subsection{Ablations and Analysis}
\label{sec:exp:ablations}

\begin{figure}[t]
\centering
\includegraphics[width=0.9\columnwidth]{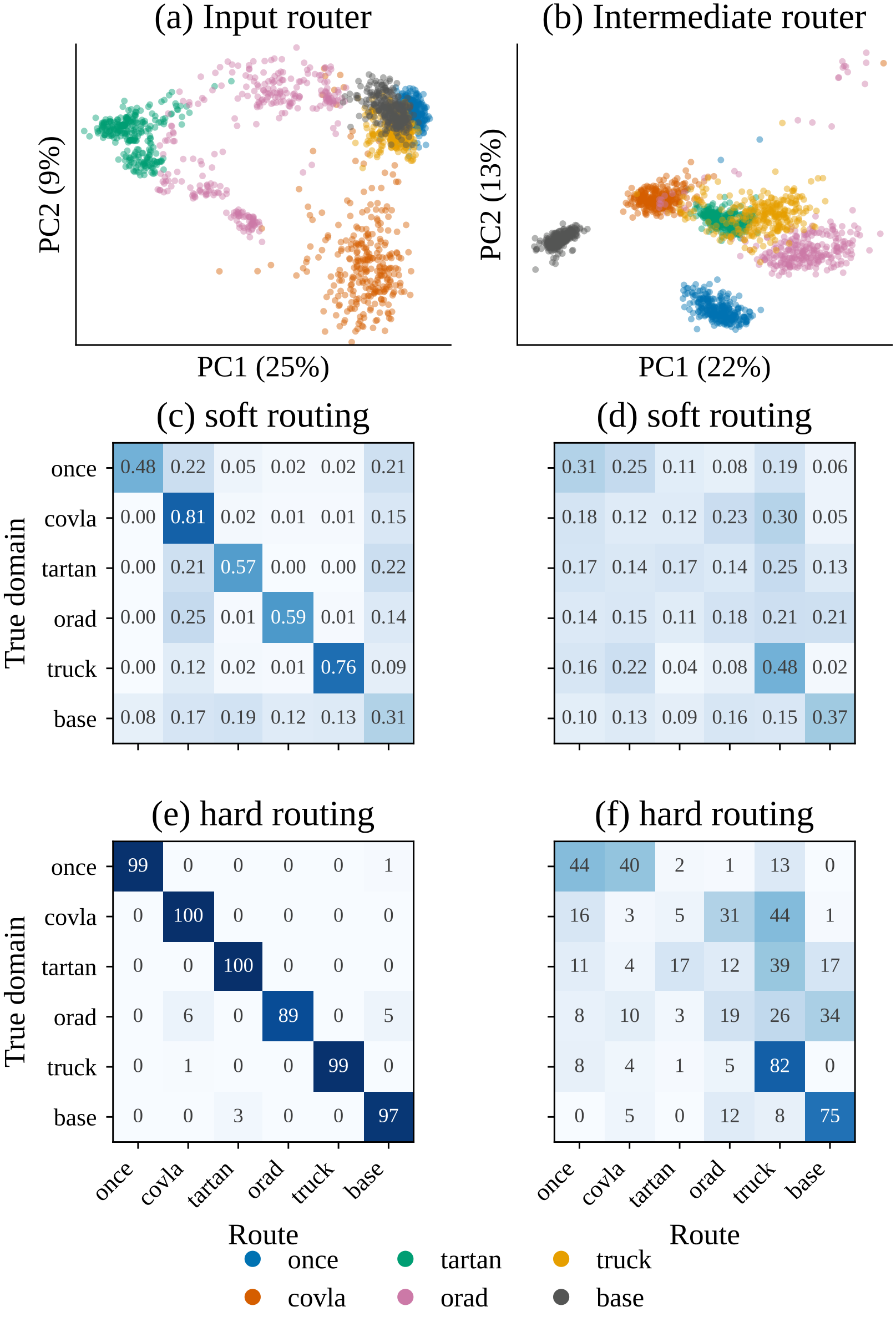}
\caption{Router behavior. Each column is one router, with rows showing its
input-feature PCA (top), soft routing (middle), and hard routing (bottom),
the confusion matrices row-normalized by true domain. The input router (left)
recognizes the domain and picks the corresponding route $97\%$ of the time, while the
intermediate router (right) routes to the correct domain only $40\%$ of the time, indicating that the dynamics experts are
interchangeable.}
\label{fig:router}
\end{figure}

\textbf{Router behavior.}
The two routers (Sec.~\ref{sec:method:routing}) answer different questions
(Fig.~\ref{fig:router}). Once each feature dimension is standardized, both see
inputs that separate the domains. The input router does domain recognition and
routes $97\%$ of clips to its own expert. The intermediate router sees equally
separable features but picks the own domain only $40\%$ of the time: the
flow-matching dynamics experts are interchangeable. It mixes experts across domains
and makes top-k routing effective (Table~\ref{tab:main}).

\begin{figure}[tb]
\centering
\includegraphics[width=0.85\columnwidth]{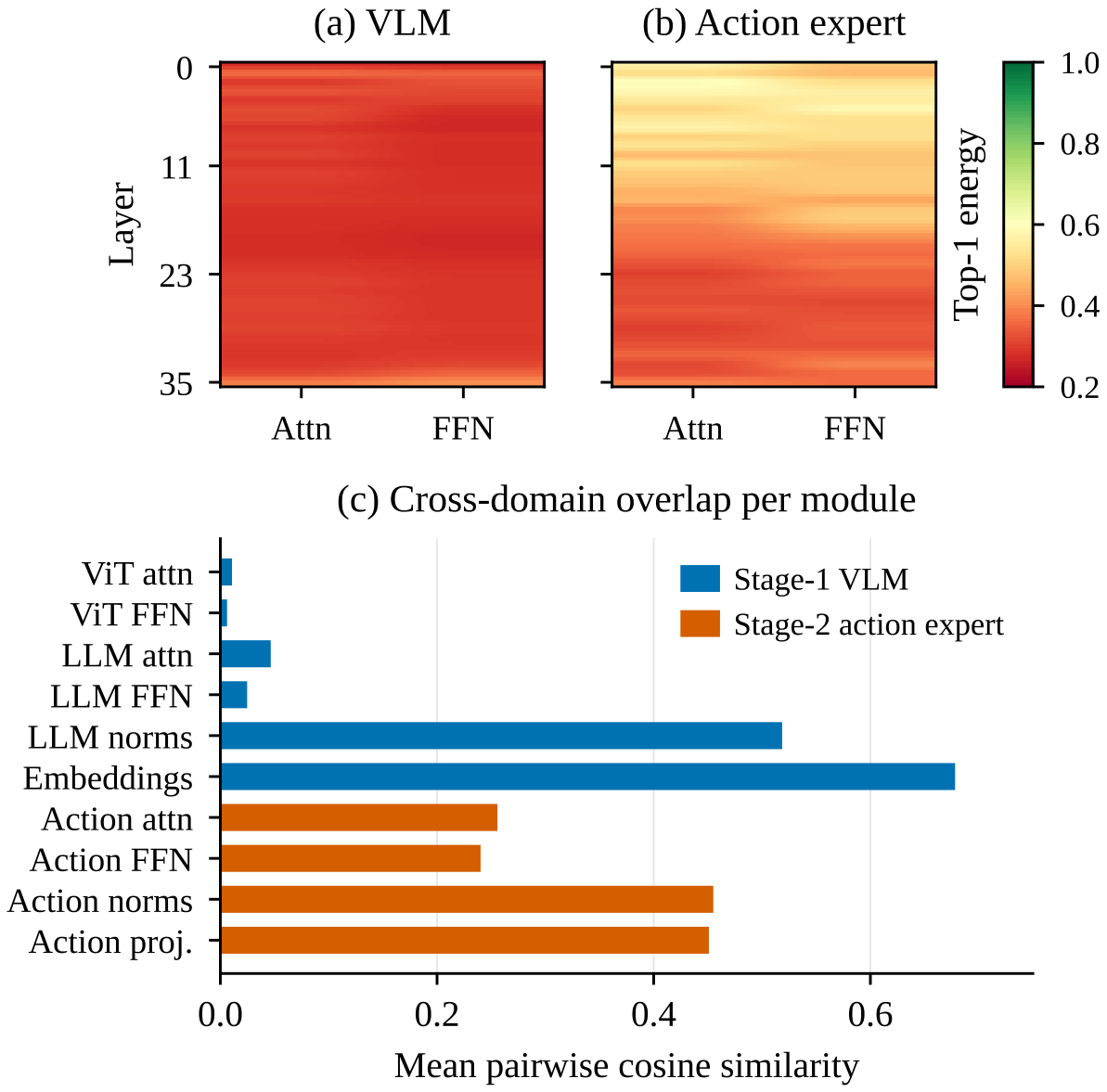}
\caption{Task-vector spectrum per module. (a,b)~Per-layer top-1 energy of
attention and FFN in the VLM and the action expert, red marks the low,
near-orthogonal band. (c)~Mean pairwise cosine similarity of the domain deltas per module:
attention and FFN near zero (route), embeddings and norms high (merge).}
\label{fig:spectrum}
\end{figure}

\textbf{Route vs.\ merge.}
This ablation asks, per module, where to route and where to merge. For each
weight matrix we stack the $N$ per-domain deltas and measure the top-1 energy
$\sigma_1^2/\sum_k\sigma_k^2$ (Fig.~\ref{fig:spectrum}a,b) and the mean pairwise
cosine of the deltas (Fig.~\ref{fig:spectrum}c). Attention and FFN have low top-1 energy (below $0.30$
in the VLM) and near-zero cosine, so the domains move them in almost orthogonal
directions and we route them as low-rank cores. Token embeddings and norms are
the opposite, with cosine up to $0.68$, so we merge them into one shared copy.
This sets the factorization scope of Sec.~\ref{sec:method:pef}: route attention
and FFN, merge the rest.

\textbf{Routing granularity.}
A clip is a single trajectory, so \method{} routes all cores once per clip. The
finer routers used in LLM MoEs do not fit this setting. A per-layer variant places
one MLP router at each of the 36 LLM layers, and a per-token variant routes every
token at every layer. Both are made for end-to-end training, where the routers and
experts are learned together at full cost, not for post-hoc composition of frozen
cores. 
% With the same cores and soft-oracle labels, per-layer routing raises the
% six-domain mean \minade{} from $1.0157$ to $1.1589$ ($+14.1\%$) and per-token
% routing to $1.1015$ ($+8.4\%$), and the loss tracks how often the fine-grained
% decisions match the clip route ($65\%$ and $67\%$ coherence, respectively).
With the same cores and soft-oracle labels, per-layer routing raises the 
six-domain mean \minade{} from $1.0157$ to $1.1589$ ($+14.1\%$) and 
per-token routing to $1.1015$ ($+8.4\%$). In both cases, the fine-grained 
routing decisions agree with the clip-level route only $65\%$ and $67\%$ of 
the time respectively, confirming that sub-clip switching adds noise rather 
than signal.

\textbf{Reasoning retention.}
\method{} keeps the base model's driving reasoning. On LingoQA~\cite{lingoqa},
its zero-shot Lingo-Judge accuracy is $73.8\%$ with top-1 routing and $73.0\%$
with top-2, matching the base model at $73.2\%$. LingoQA has no ego trajectory,
so the input router sends $89\%$ of questions to the base route and the base
reasoning stays intact. Full per-domain finetuning drops this score by over $25$
points on the most out-of-distribution domain.

%% file: sections/05_conclusion.tex
\section{Conclusion}
\label{sec:conclusion}

We presented \method{}, a post-hoc mixture-of-experts VLA that consolidates
independently finetuned per-domain driving models into one multi-domain VLA
from their weights and a small proxy set, using Post-hoc Expert Factorization to split
the finetuning deltas into a frozen shared skill basis and per-domain low-rank
cores that two cooperating clip-level routers mix at inference. Across six
heterogeneous domains, \method{} approaches the per-domain experts and
outperforms merging, routing, and data-pooling baselines at near
single-model cost, with cheap new-domain opt-in and compositional
generalization to unseen domains. In future work, we plan to scale \method{} to
more domains and base VLA models.